\documentclass{easychair}

\usepackage[round]{natbib}

\usepackage[figure,linesnumbered,boxed,norelsize]{algorithm2e}
\usepackage{fancyvrb}

\usepackage{makeidx}
\makeindex

\usepackage{bibentry}
\nobibliography*

\newcommand{\xf}[1]{Figure~\ref{#1}}

\newcommand{\xs}[1]{Section~\ref{#1}}

\newcommand{\gipsy}{{GIPSY\index{GIPSY}}}

\newcommand{\cpp}{{C++\index{C++}}}

\newcommand{\java}{{Java\index{Java}}}

\newcommand{\file}[1]{\url{#1}\index{Files!#1}}
\newcommand{\tool}[1]{\texttt{#1}\index{Tools!#1}}

\newcommand{\api}[1]{\texttt{#1}\index{API!#1}}

\newcommand{\marf}[0]{MARF\index{MARF}\index{Frameworks!MARF}\index{Libraries!MARF}}

\newcommand{\lucidL}[1]{{$\mathit{Lucid}$}($L$) }

		{}

\def\myvert{\raise 2.27pt \hbox{\vrule depth 0pt height 8pt width 0.2mm}}
\def\myarrow{\hspace*{0.43mm}%
             \raise 2.29pt\hbox{\vrule depth 0pt height 8pt width 0.16mm}%
             \hspace*{-0.32mm}%
             $\longrightarrow$
             \ %
             }

\lstdefinestyle{codeStyle}
{
        language=Java,
        frame=single,  
        basicstyle=\footnotesize,
        captionpos=b,
        showstringspaces=false,
        showspaces=false,
        extendedchars=true,
        linewidth=1\linewidth,
        breaklines=true,
        float=phtb  
}

\makeindex

\begin{document}

\title{OCT Segmentation Survey and Summary Reviews and a Novel 3D Segmentation Algorithm and
a Proof of Concept Implementation}

\author{Serguei A. Mokhov$^{1,2}$\\
$^{1}$Tsinghua University, Beijing, China\\
$^{2}$Concordia University, Montreal, QC, Canada\\
Yankui Sun$^{1}$\\
$^{1}$Tsinghua University, Beijing, China}

\date{Revision: 1.34  : Date: 2012/06/09 01:57:54  }

\begin{titlepage}
\maketitle

\begin{abstract}
We overview the existing OCT work, especially the practical aspects of it.
We create a novel algorithm for 3D OCT segmentation
with the goals of speed and/or accuracy while remaining flexible in the
design and implementation for future extensions and improvements.
The document at this point is a running draft being
iteratively ``developed'' as a progress report as the work and survey advance.
It contains the review and summarization of select OCT works,
the design and implementation of the OCTMARF experimentation application
and some results.
\end{abstract}

\thispagestyle{empty}
\end{titlepage}

\clearpage

\tableofcontents
\listoffigures

\clearpage

\section{Objectives and Motivation}

Much of the work was done on 2D-segmentation of retinal layers;
primarily by processing images one slice at a time.
Disadvantages of this kind of method include: 

\begin{enumerate}
	\item 
	relationship between images is not used;
	\item 
	processing time is expensive for OCT-volume data
\end{enumerate}

As a result, there is more and more 3D segmentation research.
Thus, the objectives of this work include the
review the related work and a proposal of new scheme to segment retinal scans.
do some PoC experiments.

\section{Literature Review}

This is an overview of the prominent papers in the OCT to comprise the literature review.
Some of the the specific papers reviewed
are
\cite{auto-macular-segmentation-oct-gradient-2010,%
auto-segmentation-macular-oct-performance-2011,%
auto-macular-segmentation-oct-2009,%
segmentation-3d-retinas-oct-2007,%
oct-auto-ophthalmic-2010}, and others
based on their practical aspects. What follows are
either citations or their summary descriptions in
a common algorithmic form with the purpose of practical
realization in a framework or comparative studies
later on.

\subsection{Background Review Papers}

\begin{itemize}
\item
	The original invention of OCT is documented in \cite{oct-1991}.

\item
	\bibentry{retinal-oct-2008}

\item
	One of the recent student survey papers on the OCT -- \cite{oct-auto-ophthalmic-2010};
	in 8 IEEE pages it discusses the recent (prior and including 2010) approaches to the OCT image processing.
	The authors first discuss in detail the OCT technology from image acquisition hardware and software perspectives, applications,
	visualization, followed by the survey of the segmentation literature covering de-noising
	(speckle noise removal, including media filters, wavelets, ZAP, nonlinear,
	diffusion filters, 7x5 mean filter) and analysis (RFE high reflectivity, Markov boundary models,
	coherence matrix, peak and valley analysis in A-scans, adaptive thresholding, various multi-step
	approaches, active contours, optimal 3D graph search, random contour based analysis,
	support vector machines (SVM), deformable fluid-filled models).
	It is very well illustrated.
	The authors touch on some of the papers we have also briefly reviewed below, such as
	\cite{segmentation-3d-retinas-oct-2007,auto-macular-segmentation-oct-2009}.
	We also have on our list of references intersecting with theirs 
	\cite{oct-1991,%
	retinal-thickness-oct-markov-2001,%
	macular-segm-oct-2005,%
	auto-retinal-layers-oct-2005,%
	quantitative-retinal-features-oct-2007,%
	retinal-oct-2008}, which is
	maybe easier to refer to \citet{oct-auto-ophthalmic-2010}'s summary instead
	for.
	The authors also highlight the areas that still need more work in the OCT image
	processing back in 2010, in particular in automation, and some errors in algorithms
	proposed by various authors not being able to cope with either healthy or diseased eyes,
	problems introduced by de-noising (decrease in speed and information loss)
	Some of their concerns have been answered to a degree by newer works published
	later in 2010 and 2011 and some of which we reviewed in
	\cite{auto-macular-segmentation-oct-gradient-2010,auto-segmentation-macular-oct-performance-2011}.
	\citet{oct-auto-ophthalmic-2010} themselves do not offer newer algorithms in that
	paper to solve some of the highlighted problems.
\end{itemize}

\subsection{2D Segmentation Papers}

This is a representative list of the 2D segmentation works that
we have not reviewed in much detail in here, but provide for
reference to the reader. Additional summary reviews may be
added at a later date.

\begin{itemize}
\item
	\bibentry{retinal-thickness-oct-markov-2001}

\item
	\bibentry{auto-retinal-layers-oct-2005}

\item
	\bibentry{macular-segm-oct-2005}

\item
	\bibentry{retinal-thickness-map-oct-2005}

\item
	\bibentry{posterior-retina-analysis-oct-2007}

\item
	\bibentry{quantitative-retinal-features-oct-2007}

\item
	\bibentry{fuzzy-cuts-cvpr-2009}

\item
	\bibentry{improving-img-segm-grading-oct-2010}

\item
	\bibentry{retina-layers-oct-active-contour-2011}
\end{itemize}

\subsection{3D Segmentation Papers}

This section lists majority of the review summaries in the 3D segmentation
area.

\begin{itemize}
\item
	\citet{segmentation-3d-retinas-oct-2007} (a follow up work on the author's
	first attempt in \cite{vector-machine-retina-oct-2007}) propose the use of the 
	support vector machines (SVMs) and machine learning to do the segmentation of a 3D
	volume. The approach is interactive and involves comprehensive
	visualization. A 3D volume is constructed and the clinicians
	interact with it by placing control points at the layers of interest
	and then the system interpolates in between.
	These become them the input to the SVN to learn and further segment
	retina scans in 3D. In this latest approach of the authors a multi-resolution
	imaging is used of the same area of interest in order to combat speckle noise
	and improve overall precision. The run-times are approx. within the range
	15-30 minutes depending on the hardware. This is a good example of a machine
	learning segmentation, that claims to do better than neural networks, etc.
	but it also requires interactive user sessions before it can classify reliably
	the retina layers.
	The summary is in \xf{algo:segmentation-3d-retinas-oct-2007}.

\begin{algorithm}[hptb]
\SetAlgoLined
Load 3D OCT scan and visualize\;
\tcp{Interactive step by clinicians}
Specify vertices of interest for each layer\;
\Begin
{
	Location of the voxel in the 3D volume: $p_{i}(x_{i},y_{i},z_{i})$\;
	Scalar intensity at $p_{i}$: $f_{i}$\;
	$p_{i}$'s neighbors: $N = \{i \pm 1, j \pm 1, k \pm 1\}$\;
	Mean intensity value of $p_{i}$'s neighbors: \[\overline{f_{i}}=\frac{1}{|N|}\sum_{n \in N}{f_{n}}\]\;
	Variance around $p_{i}$: \[\sigma_{i}^{2}=\frac{1}{|N|}\sum_{n \in N}{\left[f_{n}-\overline{f_{i}}\right]^{2}}\]\;
	Gradient around $p_{i}$: \[\nabla\overline{f_{i}}=\frac{1}{|N|+1}\left\{\nabla f_{i}+\sum_{n \in N}\nabla{f_{n}}\right\}\]\;
}
Interpolate\;
Perform 4-fold multi-resolution image split\;
Interpolate the layer points for each resolution\;
Machine-learn the points using kernel-based SVN using
features such as gradient, intensity, variance, and
spacial coordinates\;
Classify and measure layer thickness\;
\caption{Synthesized from \citet{segmentation-3d-retinas-oct-2007}}
\label{algo:segmentation-3d-retinas-oct-2007}
\end{algorithm}

\item
	\bibentry{aut-seg-rnfl-oct-2009}

\item
	From roughly the same people
	\cite{retina-layers-3d-oct-graph-search}, and
	\cite{retina-layers-auto-3d-oct-seg-2009} on 
	3D OCT.

\item
	Review for practical aspects:
	\cite{auto-macular-segmentation-oct-2009}.
	The authors automatically cover 2 layers:
	ILM and RPE, but they do it across the entire
	3D volume. It seems they are pretty fast in most
	of their iterative algorithms where the number of
	iterations is configurable and serves as a trade off
	between quality of the segmentation and speed. Claimed
	speeds on a PC range between $16s$ to $21s$. The ILM
	classification in particular is worth building upon.
	They use signal intensity variation based segmentation;
	no denoising on both healthy and diseased retinas.
	The algorithmic summary is in \xf{algo:auto-macular-segmentation-oct-2009-rpe}
	and \xf{algo:auto-macular-segmentation-oct-2009-ilm}.
	Their RPE and ILM can be done in parallel, so speeding up
	the overall performance on parallel and distributed architectures.

\begin{algorithm}[hptb]
\tcp{Preprocessing is not done, but can be done if wanted}
\tcp{RPE identification}
\Begin
{
	\tcp{$x$ -- width, $z$ -- height, $y$ -- depth in slices (B-scan)}
	Determine coordinates of max. intensity $\max(I_{x,y}(z))$
	pixels $z_{\max(I)}(x,y)$ across volume on $(x,y)$\;
		
	Obtain 2D RPE position matrix $z_{\max(I)}(x,y)=z_{rpe1}(x,y)$\;
	
	\tcp{Takes care of the speckle noise from RNFL}
	Using Otsu method, binarize to get erroneous pixels mask $B_{rpe1}(x,y)$ for RPE position matrix in $(x,y)$
	using top-hat filtering using structuring element $5\times5$ pixels\;
	
	Replace erroneous pixes with values from nearest neighbors in$(x,y)$\;
	\Begin
	{
		\tcp{Set expected erroneous pixels' value to $NaN$}
		\If{$B_{rpe1}(x,y)$ = 0}{$z_{rpe1}(x,y) = NaN$\;}
		$Z_{rpe1}(x,y) = $\ForAll{$z_{rpe1}(x,y) = NaN$}
		{
			$z_{rpe1}(x,y) = $ based on the nearest neighbor value\;
			Smooth by moving window median filter $30(x) \times 2(y)$ pixels\;
		}
	}
	
	\tcp{30 initial pixels}
	$P=\{I_{x,y}(z), z \in [Z_{rpe1}(x,y)-10,Z_{rpe1}(x,y)+20]\}$\;
	\For{selected number iterations}
	{
		Extract $P$ pixels around RPE estimation from the original 3D volume $(x,y,x)$\;
		Update RPE position based on the maximum intensity\;
		Smooth by moving window median filter $40(x) \times 2(y)$ pixels\;
		Reduce amount of pixels $P$ around RPE to 20, and then 10
		to do $P=\{I_{x,y}(z), z \in [Z_{rpe2}(x,y)-10,Z_{rpe2}(x,y)+10]\}$
		and $P=\{I_{x,y}(z), z \in [Z_{rpe3}(x,y)-5,Z_{rpe3}(x,y)+5]\}$\;
	}
	Smooth by moving window median filter $20(x) \times 2(y)$ pixels\;
}
\caption{Synthesized RPE Detection from \citet{auto-macular-segmentation-oct-2009}}
\label{algo:auto-macular-segmentation-oct-2009-rpe}
\end{algorithm}

\begin{algorithm}[hptb]
\tcp{ILM identification}
\Begin
{
	Calculate threshold value and binarize each B-scan slice\;
	\Begin
	{
		\ForEach{B-Scan in data cube}
		{
			Extract first 5 depth (assumed noisy) pixels
			$\{Noise_{y}(x,z), y \in [1,N], x \in [1,M], z \in [1,5]\}$\;
			Evaluate $5 \times M$ pixels\;

			\tcp{$Bnoise_{y}(x,z)$ is binarized $Noise_{y}(x,z)$ as each Threshold is different}
			Threshold = $0.5\%$ of pixels set to 0 after binarization
			$\{\sum_{x=1}^{M}{\sum_{z=1}^{5}{Bnoise_{y}(x,z) \leq 0.005 \times 5 \times M}}\}$\;
		}
	}

	Do first ILM estimation via depth position of the first zero value of each binarized A-scan\;
	\For{number of iterations}
	{
		Extract 45 pixels around estimated ILM and reprocess to remove erroneous ILM pixels\;
		\Begin
		{
			$\{S_{y}(x,z), y \in [1,N], x \in [1,M], z \in [Z_{ilm}(x,y)-15,Z_{ilm}(x,y)+30]\}$\;
		}
		Smooth by moving window medial filters $1(x) \times 25(y)$ and $25(y) \times 1(x)$\;
		Binarize using intensity with the same threshold\;
		Re-estimate ILM position based on the first zero depth value of A-scans\;
		
		Repeat with 30 pixels, $[Z_{ilm2}(x,y)-3,Z_{ilm2}(x,y)+27]$, and filter $10(x) \times 1(y)$\;
	}
}
\caption{Synthesized ILM Detection from \citet{auto-macular-segmentation-oct-2009}}
\label{algo:auto-macular-segmentation-oct-2009-ilm}
\end{algorithm}

\item
	\bibentry{robust-retinal-segmentation-statistical-2010}

\item
	\bibentry{retinal-thickness-map-oct-2005}

\item
	This work of \citet{auto-macular-segmentation-oct-gradient-2010}
	uses {\bf gradient information} to get 9 layers; they are fast
	and thorough, but the paper omits a few algorithm details. The first
	author is from Topcon, the maker of the OCT 1000 imaging hardware.
	Subjects -- 38 (19 healthy and 19 glaucoma), including live and dead tissue.
	The results are 	compared to 4 experts' manual segmentation.
	
	The parts of algorithm include two steps -- Canny edge detection
	(customized with 3 thresholds, graph-based node cost assignment,
	dynamic programming (during the shortest path search). The gradient
	aspect seems useful and fast.
	
	The summary in the algorithm-like notation is in
	\xf{algo:auto-macular-segmentation-oct-gradient-2010}.

\SetAlFnt{\scriptsize}

\begin{algorithm}[hptb]
\SetAlgoLined
\KwData{480x512x128 voxels}
\KwResult{9 segmented boundaries in 16 seconds}
\tcp{2-step segmentation based on gradient:}

\lForEach{408x512 A-scan $a$ of the 3D OCT Volume}
{
	Obtain gradient information\;
	\Begin
	{
		Local customized Canny main edge result with 3 thresholds\;
		\Begin
		{
			$C(i, j) = w_{1} \cdot Canny(i, j) + w_{2} \cdot Axial(i, j) + w_{3} \cdot Others(i, j)$\;
		}
	
		\BlankLine
		\tcp{To interpolate missing/weak gradient info (over blood vessels, etc)}
		Global axial intension gradient\;
	
		Apply the shortest path search to complete and optimize detection\;
	
		\Begin
		{
			Dynamic programming to optimize the result
			\[
				t(i,j) = \left\{
				\begin{array}{l l}
					\infty & \quad j<1, j>m\\
					C(i,j) & \quad i=n\\
					\min_{m=j-2:j+2}{(t(i-1,m))}+C(i,j) & \quad \text{otherwise}\\
				\end{array} \right.\;
			\]
		}
	}
}

\tcp{Nine boundary detection}
\lForEach{B-scan $b$ of the 3D OCT Volume}
{
	\Begin
	{
		Preprocess the OCT image with a customized cross-correlation based alignment
		algorithm to realign the A-scans\;
		
		Detect the ILM and IS/OS boundaries (the shortest path
		search maps of the other boundaries can be restricted to successively smaller
		search areas)\;
		
		Add features {\em edge direction} and {\em pixel intensity} as the
		additional cost to the search graph depended on the boundary of interest.
		The edge direction is only considered as black-to-white or white-to-black\;
		
		Align the image further the IS/OS boundary\;

		Detect the OS/RPE and BM/Choroid boundaries\;
		\Begin
		{
			Accurately detect the OS/RPE and BM/Choroid via a smaller kernel
			size applied in the Canny edge detector to increase the axial detection
			sensitivity\;
			
			The search graph is constructed using only Canny edge and axial
			intensity gradient strength, and the search area is limited below
			the IS/OS boundary\;
		}

		Detect IPL/INL and NFL/GCL\;
		Detect GCL/IPL within the NFL/GCL and IPL/INL boundaries\;
		\Begin
		{
			Apply the Canny edge detector again within that area to extract the GCL/IPL boundary\;
		}
		
		Detect the INL/OPL, the dark-to-bright edge between the IPL and the outer nuclear layer\;
		
		Detect the ELM similarly via the dark-to-bright edge\;

		\tcp{Although the detection of a single boundary may utilize the other two pre-detected
		neighboring boundaries to limit the shortest path search area, the intra-retinal boundaries are
		allowed to overlap their neighboring boundaries.}

		Smooth intra-retinal boundaries (NFL/GCL, GCL/IPL, IPL/INL, ELM, OS/RPE, BM/Choroid)
		using a polynomial curve-fitting based technique\;
	}
}

\BlankLine
Apply additional smoothing across frames for 3D volumes\;
Filter the boundaries for each A-scan location with a 1D Gaussian kernel across B-scans\;

\caption{High-Level Algorithm for Fast 9-Layer 3D Segmentation by \citet{auto-macular-segmentation-oct-gradient-2010}}
\label{algo:auto-macular-segmentation-oct-gradient-2010}
\end{algorithm}

\item
	\bibentry{oct-auto-ophthalmic-diagnosis-2010}

\item
	The result \citet{auto-segmentation-macular-oct-performance-2011}
	of is also on 3D using similar equipment to
	\citet{auto-macular-segmentation-oct-gradient-2010} and 8 layers
	using a different approach. The paper is very detailed in the
	algorithms used.
	The authors use active contours,
	Markov random fields, Kalman filters.
	
	The authors don't seem to claim speed. Denoising is in play. Very detailed literature
	review. They have a lot larger database of 700 images (but only on healthy
	subjects) and test their approach
	from two separate imaging hardware types: Topcon's 3D OCT 1000 and Spectralis HRA+OCT(Heiderlberg)
	and 5 experts vs. 4 in \citet{auto-macular-segmentation-oct-gradient-2010}.

\begin{algorithm}[hptb]
\SetAlgoLined
Preprocessing\;
HRC Detection\;
ILM Localization\;
Photoreceptor (IS-OS) Segmentation\;
Alignment/Clivus Detection\;
Inner Layers Segmentation\;
\caption{Overall algorithm by \citet{auto-segmentation-macular-oct-performance-2011}}
\label{algo:auto-segmentation-macular-oct-performance-2011}
\end{algorithm}

\begin{algorithm}[hptb]
\SetAlgoLined
\Begin
{
	Crop image to OCT data\;
	Normalize $[0..1]$\;
	Apply non-linear diffusion filter \cite{non-linear-diffusion-filters-1998}\;
}
\caption{Synthesized Preprocessing from \citet{auto-segmentation-macular-oct-performance-2011}}
\label{algo:auto-segmentation-macular-oct-performance-2011-prep}
\end{algorithm}

\begin{algorithm}[hptb]
\SetAlgoLined
\Begin
{
	\tcp{W is the image width}
	$y_0 = W/2$\;
	Select pixel $(x_0,y_0)$ as max response to avg. filter\;
	Estimate HRC thickness $T_{HRC}$ via profile analysis\;
	Detect HRC median line\;
	\Begin
	{
		(1) Vertically smooth image $S(x,y) = $ apply 1D Gaussian filter with
		std. deviation $\sigma=T_{HRC}/2$\;
		(2) \tcp{Regular median line regardless noise and blood vessels}
		$\alpha = 0.9$\;
		From $(x_0,y_0)$ column-wise in both directions deduct median line
		iteratively from max. output of the recursive low-pass filter:
		$\max(C(x,y)=(1-\alpha)S(x,y)+\alpha C(x,y \pm 1))$\;
		(3) Localize HRC contours\;
		\Begin
		{
			Perform $k$-means classification ($k=3$)\;
			From the median line initialize active contour curve
			$X(s)=[x(s),y(s)], s \in [0,1]$\;
			\Begin
			{
				\tcp{The active contour is used for regularization purposes}
				Set $\alpha$ and $\beta$ for contour tension and rigidity\;
				Compute data fidelity $E_{Image}(X(s))$ as a spatial diffusion of the gradient of an edge map\;
				Move $X$ within the image's spatial domain to minimize energy $E(X)$:
				$E(X)=\int\limits_0^1\frac{1}{2}\left(\alpha|X'(s)|^2 + \beta|X''(s)|^2 + E_{Image}(X(s))\right)ds$\;
			}
		}
	}
}
\caption{Synthesized HRC Detection from \citet{auto-segmentation-macular-oct-performance-2011}}
\label{algo:auto-segmentation-macular-oct-performance-2011-hrc}
\end{algorithm}

\begin{algorithm}[hptb]
\SetAlgoLined
\Begin
{
	Re-use $k$-means classification ($k=3$)\;
	\lForEach{Image column $c$ scanned top-bottom}
	{
		\Begin
		{
			\tcp{Pick a boundary pixel $p(c,y)$ if}
			\Switch{Pixel}
			{
				\uCase{$p$ = first pixel of $k=3$}{}
				\uCase{$p$ above HRC}{}
				\Case{$p$ = first pixel of $k=2$}{Pick $p(c,y)$\;}
				\Other{Don't pick $p(c,y)$}
			}
		}
		Select 2 highest gradient $p_l$ and $p_r$ from the left and right parts of the image\;
		Apply edge-tracking algorithm maximizing local gradient starting from $p_l$ and $p_r$\;
		Merge resulting curves maximizing the mean gradient\;
		Refine ILM boundary by active contour\;
		Identify foveola $F(x_F,y_F)$ where $x$ is maximal on ILM\;
	}
}
\caption{Synthesized ILM Localization from \citet{auto-segmentation-macular-oct-performance-2011}}
\label{algo:auto-segmentation-macular-oct-performance-2011-ilm}
\end{algorithm}

\begin{algorithm}[hptb]
\SetAlgoLined
\Begin
{
	Apply a peak detector on the image area WRT HRC\;
	Detect and label maxima in each column to form peak lines\;
	Init IS/OS junction extraction by selecting the peak line
	with min distance to HRC\;
	Iterate over other peak lines WRT the current curve\;
	\Begin
	{
		Similar mean distance to HRC without overlap\;
		Apply active contour using detected curve to initialize
		to get a contiguous regular curve\;
	}
	Detect ONL/IS boundary\;
	\Begin
	{
		\lForEach{Column $n$}
		{
			Apply Kalman filter with a state vector $X$\;
			\Begin
			{
				Init from the peak line above IS/OS\;
				$X = (Intensity, Distance)$\;
				\tcp{identity matrix}
				$F=I$\;
				Prediction: $\hat{X}(n | n-1)=F\hat{X}(n - 1 | n - 1)$\;
				$Y(n)=$ max intensity pixel around $\hat{X}(n | n-1)$; 0 otherwise\;
				\tcp{$G(n) adjusted each iteration$}
				$G(n)=$ weighted Kalman filter gain for error correction\;
				Update: $\hat{X}(n | n)=\hat{X}(n | n-1) = G(n)(Y(n) - \hat{X}(n - 1)$\;
				Stop: $Y(n)=0$ for several iterations\;
				Regularize detective ONL/IS boundary curve by active contour\;
				\If{Image is of good quality}
				{
					Apply second Kalman filter to detect OS/RPE boundary similarly to the above\;
				}
				\Else
				{
					Apply $k$-means with $k=2$ between IS/OS and HRC\;
					\tcp{Provides first estimation around RPE-ChChap inner boundary}
					Estimate thickness RPE-ChChap thickness $T_{RPE+ChChap}$\;
					Perform local analysis of the estimated curve for $n$\;
					\Begin
					{
						OS/RPE pixel's $y$ is a first pixel of max gradient below min intensity
						below the IS/OS junction\;
						Regularize detected curve by active contour\;
					}
				}
			}
		}
	}
}
\caption{Synthesized Photoreceptor (IS-OS) Segmentation from \citet{auto-segmentation-macular-oct-performance-2011}}
\label{algo:auto-segmentation-macular-oct-performance-2011-osis}
\end{algorithm}

\begin{algorithm}[hptb]
\SetAlgoLined
\Begin
{
	Align image by vertical column translation against outside of the RPE+ChCap\;
	Define clivus as two highest points $x_{Cl},y_{Cl}$ and $x_{Cr},y_{Cr}$ on ILM\;
	Refine foveola position as $\max_{distance}(ONL/IS, OS/RPE)$\;
}
\caption{Synthesized Alignment/Clivus Detection from \citet{auto-segmentation-macular-oct-performance-2011}}
\label{algo:auto-segmentation-macular-oct-performance-2011-clivus}
\end{algorithm}

\begin{algorithm}[hptb]
\SetAlgoLined
\Begin
{
	Model noise and spatial pixel interaction using Markov Random Field (MRF)\;
	Classify with Bayesan max a posteriori (MAP) criterion to locate
	a label config maximizing $P$ based on the observed image intensity\;
	\Begin
	{
		\tcp{Probability of pixel intensities $f_s$ in class $i$}
		\tcp{$\sigma_i$ -- std. deviation}
		\tcp{$\mu_i$ -- mean}
		$P(f_s|w_s = i) = \frac{1}{\sqrt{2\pi}\sigma_i}e^{-\frac{{(f_s - \mu_i)}^2}{2\sigma^{2}_{i}}}$\;
		\tcp{Potts model pixel interactions $\phi(w_s,w_t)$ in 8-connected sites $s$ and $t$}
		\If{$w_s = i$}
		{
			$\mu_{w_s}=\mu_i$\;
			$\sigma_{w_s}=\sigma_i$\;
		}
		$\beta$ empirically set to balance image data term and regulation term in $U$\;
		\tcp{Energy function}
		\[
		U(w|f)=\sum_{s}\left[\frac{(f_s - \mu_{w_s})^2}{2\sigma^{2}_{w_s}} + \ln(\sqrt{2\pi}\sigma_{w_s})\right] + \beta \sum_{s,t}\phi(w_s,w_t)
		\]
		Run unsupervised $k$-means to init labels and initialize noise params $(\mu_i,\sigma_i)$\;
		Minimize $U$ using Simulated Annealing repeatedly on the original, but labeled image\;
		Determine active contours from the labeled image\;
		Detect RNFL\;
		\Begin
		{
			Apply $k$-means ($k=2$) to the image top around ILM\;
			Select clusters with higher mean intensity\;
			Deduce RNFL contours on both sides of foveola with active contours\;
		}
		Detect ONL\;
		\Begin
		{
			$k=3$\;
			$\beta = 2$\;
			Apply $k$-means/MRF segmentation to the foveal region
			constrained by clivus and ILM|RNFL and ONL/IS\;
			Deduce the ONL boundary from the labeled image\;
			Regularize with active contours\;
		}
		Detect INL\;
		\Begin
		{
			$k=2$\;
			$\beta = 5$\;
			Apply the above method on the right and left sides from foveola\;
			Estimate location of searched boundary as $(y_{Cl},y_{Cl}/2,y_{Cr}/2,y_{Cr})$\;
			Force the curve to pass at $y_F$ between ILM and OPL/ONL\;
			Linear interconnect the 5 points above as the first estimate\;
			\lForEach{Column}
			{
				\Begin
				{
					Transition pixels between labels 1 and 2 and close to the first
					estimate are marked as the new boundary\;
					Initialize the new boundary estimate as active contour\;
				}
				\Begin
				{
					Traverse column down-up\;
					Apply the similar process between the found curve and OPL/ONL\;
				}
			}
		}
	}
}
\caption{Synthesized Inner Layers Segmentation from \citet{auto-segmentation-macular-oct-performance-2011}}
\label{algo:auto-segmentation-macular-oct-performance-2011-inner}
\end{algorithm}

\end{itemize}

The methods in
\cite{auto-macular-segmentation-oct-2009,auto-macular-segmentation-oct-gradient-2010}
may be more practical.

\section{The Work of Sun et al. at Tsinghua}

\subsection{Previous Results}
\label{sect:sun-previous-results}

Sun et al. covered various aspects of the OCT image processing and
classification using earth mover distance, SVMs, providing for detection
of the nacre's layer thickness and other works,
e.g. \cite{oct-classification-earth-mover-2009,auto-thickness-nacre-oct-2010}.
The most recent work in a 3D algorithm for segmentation (submitted)
is by Sun and Zhang that explores position determination for ILM,
RPE, and IS/OS in a 3D volume in an efficient and robust manner by determining
intensity change on both sides of the boundary simultaneously and then
smoothen the recovered surface using 3D intensity difference \citet{sun-3d-oct-3d-diff-op}.
Some segmentation results are shown in \xf{fig:sun-previous-oct-results},
and more are within the cited work.

\begin{figure}%
	\includegraphics[width=\columnwidth]{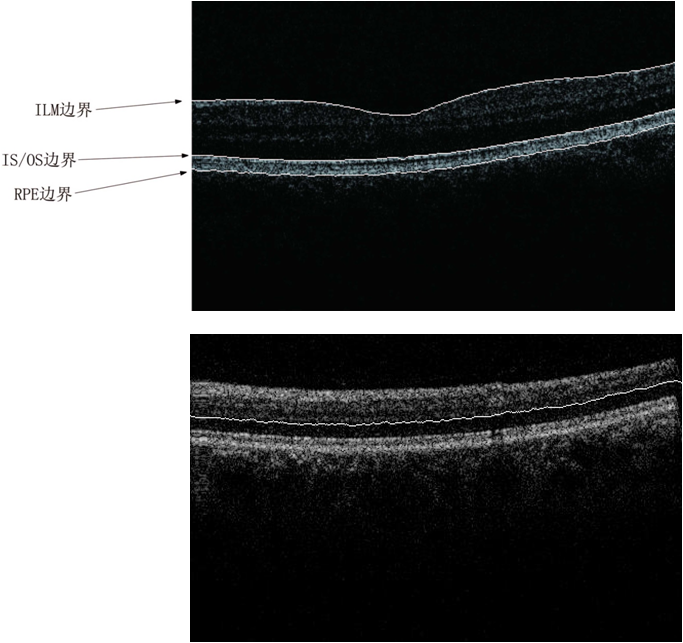}%
	\caption{Sun et al. Some Segmentation Results}%
	\label{fig:sun-previous-oct-results}%
\end{figure}

\subsection{Data Description}
\label{sect:data-description}

Data format description from Sun et al. with the clinical OCT data provided
by Shenzhen MOPTIM  Imaging Technique Co., Ltd.

\begin{itemize}
\item Text files: ASCII-encoded pixel intensity data. These are A-scans, where
each line of text corresponds to a vertical scan.
There are 100 txt files. (\file{data.zip})

\item Binary images: JPEG. 100 jpg files of OCT retinal scans (\file{pic.zip})
representing 100 slides of a 3D retina scan at 480x300x100 (96dpi) resolution.
\end{itemize}

\section{New Algorithm}

This section constructs a new algorithm or a new algorithm framework derived
from the reviewed literature on OCT and the previous works of authors brought
into the OCT domain.

\begin{enumerate}
	\item
Use gradient, Canny edge detection.

	\item
{\bf (2D done)}
Apply Zhang-Suen transform to cleanse the image and skeletonize \cite{zhang-suen-transform-1984}.
Augment the algorithm to 3D covering the nearing planes and gradient information. Use
binarization before the transform per \cite{auto-macular-segmentation-oct-2009}.

	\item
Apply 3D discrete wavelet filter (dual tree) to denoise a 3D volume from
\cite{waveletsoftware-matlab}. Potentially borrowing from
\cite{tex-img-retrieval-rcwf-2005,rotation-invar-tex-img-retrieval-rcwf-2006}.

	\item
{\bf (stub)}
{\gipsy}-based \cite{gipsy} distributed evaluation of OCT images.

	\item 
Optionally re-use Neural Network (per \cite{texture-classification-oct-2007}),
2D CFE filters \cite{shivaniharidasbhat06}, and machine learning
implementation from {\marf} (some OCT papers used neural networks
to classify pixels belonging to different layers.) \cite{marf}.

	\item 
Optionally re-use Simulated Annealing from Cryptolysis \cite{cryptolysis},
as e.g. the authors of \cite{auto-segmentation-macular-oct-performance-2011}
use it, as per \xf{algo:auto-segmentation-macular-oct-performance-2011-inner}.

	\item
\citet{oct-auto-ophthalmic-2010} in their survey pointed out a work of 
Fernandez (2005) who provided a way to detect fluid-filled areas in pathological
retinas using deformable models. Separately, Miao Song in her master thesis
\cite{msong-mcthesis-2007} implemented deformable softbody simulation, where
a 3D sphere of a spring-mass system is subject to inner pressure force opposed
to by the spring stiffness and other external forces with collision detection.
A novel idea is to use Song's softbody to fill in the regions of each layer
centered at the middles of each layer and bounded by the different intensity/gradient
value pixels on the boundaries acting as collision detection points. Putting enough
simulated inner pressure to the softbody object would make it fill up the layer accurately
enough in real-time to define two layer boundaries for most layers for the whole 3D
layer in one simulation.

  \item
\citet{oct-auto-ophthalmic-2010} also mentioned an optimal 3D graph search by Garvin
et al. in 2008 -- a novel extension of this idea is to use a 3D graph with probabilities
and treat the layer detection problem as a model-checking problem with probabilities instead
that can be learned at 3D level with some initial training/learning. For this, there
is an open-source PRISM -- probabilistic model-checking tool \cite{prism}. A model is build
for each layer with typical voxels of that layer and the corresponding graph; the probabilities
are assigned based on the local and neighbors averages or medians as well as variance
and gradient. Once model is build, graph is built and checked against the model.

	\item
Export as OBJ file format as well as 3D visualization of the entire {\bf segmented} volume
in OpenGL~\cite{opengl} to enhance perception of 3D layers and ``peek'' inside the 3D retina.
This is a an easy and cost-effective way to visualize the results in 3D instead of just each
2D slice of the volume scan.
TODO: cite \tool{volview} and other related visualization work

\end{enumerate}

\begin{algorithm}[hptb]
\SetAlgoLined
\tcp{Preprocessing (optional)}
\Switch{preprocessing method}
{
	\Case{Skeletonize/Thin/Fill/Contourize}
	{
		Optionally binarize per \citet{auto-macular-segmentation-oct-2009}\;
		Use augmented 3D Zhang-Suen transform \citet{zhang-suen-transform-1984}
		to cleanse the image and skeletonize it\;
	}
	\Case{Wavelet}{Use 3D dual-tree wavelet transform filter \citet{waveletsoftware-matlab}\;}
	\Case{None}{Skip preprocessing\;}
}

\tcp{Analysis and Detection}
\Switch{detector method}
{
	\Case{ILM}
	{
		3D ILM detection per \citet{auto-macular-segmentation-oct-2009}\;
	}
	\Case{Canny}
	{
		Apply the customized Canny edge detection from \citet{auto-macular-segmentation-oct-gradient-2010}
		on the cleansed image\;
	}
	\Case{Softbody}
	{
		Expand softbody \cite{msong-mcthesis-2007} 3D ``bubble'' bounded by higher intensity pixels centered at the layers' middle
		until the expansion stops\;
		Contour of the bubble will defined the layer boundaries\;
	}
	\Case{PRISM}
	{
		\tcp{May need to learn the models}
		Build a probabilistic graph layer model\;
		Treat the problem as a model-checking problem for each pixel in 3D space with PRISM \cite{prism}\;
	}
	\Case{Simulated Annealing}
	{
	}
}

Visualize with J3D and OpenGL\;

...\;
\caption{Building Our Novel Algorithm}
\label{algo:mokhov-sun-3d-oct-1}
\end{algorithm}

\section{Design and Implementation}
\label{sect:design-impl}

\subsection{Language}

Primary language for experiments chosen for now is {\java} due to its
more formal nature, better design, memory management. Several convenient
frameworks are available for image manipulation and patter recognition, machine
learning in {\java} as well as easier file management, distributed evaluation,
web services, and others the author Mokhov is familiar with.
Any Java program can interface a {\cpp} program and vice versa via
the Java Native Interface (JNI) \cite{jni,cni}.
Most critical and interesting things can be converted to {\cpp}
IFF needed at a later date when the experiments are over.
These design decisions and implementation are based primarily on
Java 6 (1.6.29).

\subsection{Architecture}

\begin{itemize}
	\item 
\api{OCTMARFApp} -- the main application.
The tentatively called application stub for experiments has been created and
named OCTMARF. The application's overall structure at the moment of
creation was based off \api{MARFCATApp} (\cite{marfcat-arxiv,marfcat-app,marfcat-sate2010-nist})
and its predecessor \api{WriterIdentApp} \cite{marf-writer-ident,marf-writer-ident-app} and others.

	\item
The project has been organized into Java packages for various tasks and
it incorporated an old work on pattern thinning, skeletonization and 2D
feature extraction based on Zhang-Suen transform \cite{zhang-suen-transform-1984}
that is planned to be extended to 3D per the Ideas section. This code was
ported, compiles, and runs; it is found under the \api{marf.apps.oct.OCTMARF.framework.a2}
and its subpackages as of this writing. Alrready used in preliminary experiments.

	\item
See screenshot of Eclipse on the current project's architectural layout
overview in \xf{fig:eclipse-project-layout-oct}.

\begin{figure}[htpb]%
	\includegraphics[width=\columnwidth]{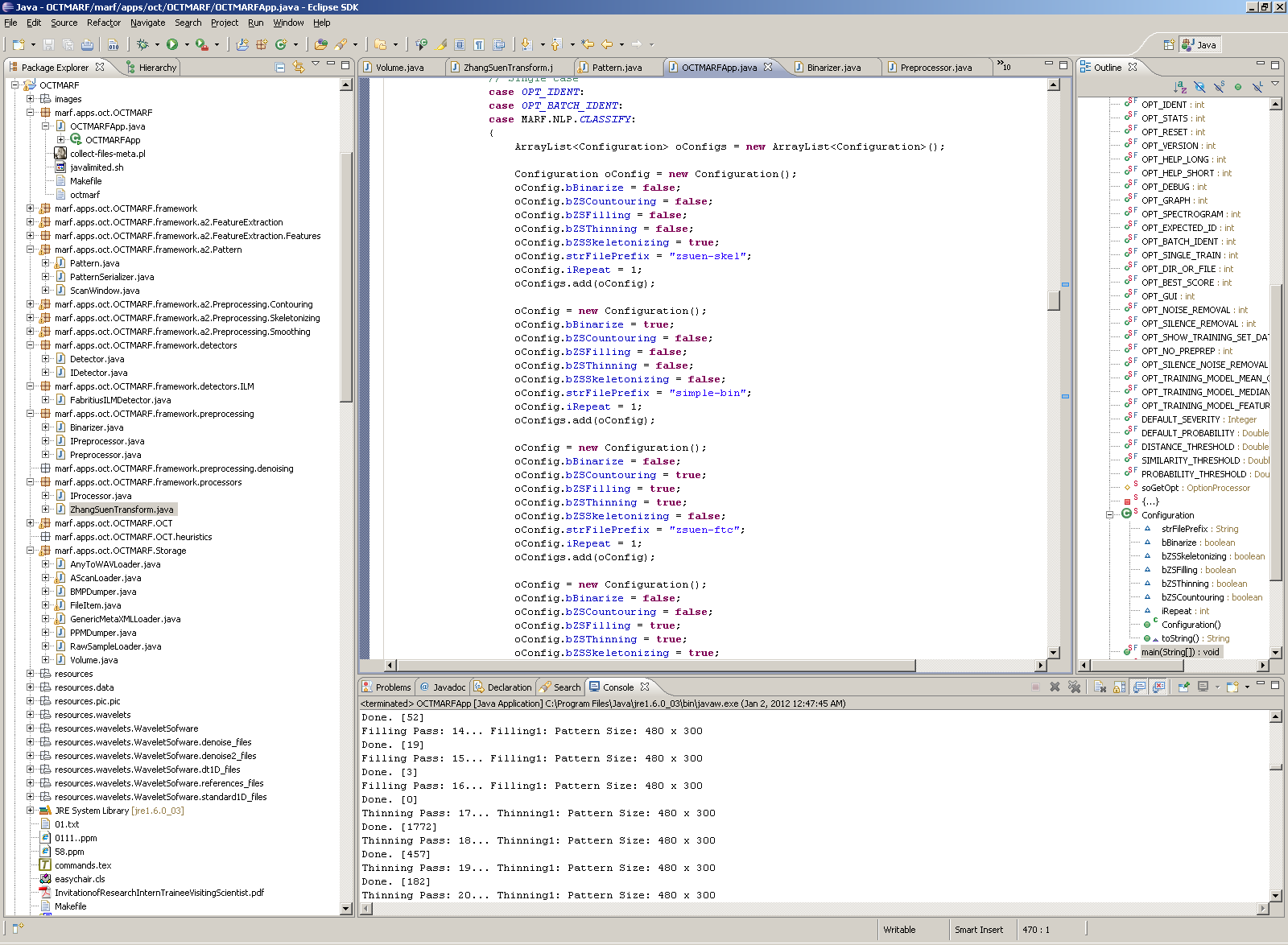}%
	\caption{Preliminary Project Layout}%
	\label{fig:eclipse-project-layout-oct}%
\end{figure}

\end{itemize}

\subsection{Data Structures}

\begin{itemize}
	\item \api{Volume} -- a class to contain the 3D OCT volume data.
	\item consists 3D arrays: \api{short}, \api{double}, \api{String} representing all the slice of a retina scan
		\begin{itemize}
			\item \api{short} -- primarily for basic loading from the .txt data and manipulation
			\item \api{double} -- for general processing, filtering, etc. as most are floating point numbers
						and many packages work with that, including {\marf}.
			\item \api{String} -- for convenience for debugging and PPM output and printing the values
		\end{itemize}
\end{itemize}

\subsection{Data Management}

\begin{itemize}
	\item \api{AScanLoader} -- a class to load the provided .txt scan data into the internal data structures.
	\item \api{PPMDumper} -- to dump .ppm image format from the internal data structures
	\item \api{BMPDumper} -- to dump .bmp image format from the internal data structures (not implemented)
	\item ...
\end{itemize}

To preliminary tests of loading, basic manipulation, and dumping are successful,
as per \xf{fig:slice58-bw-denoised-simple} and \xf{fig:slice58-bw-denoised-simple}.
These simply illustrate that the data are loaded and interpreted correctly from
the provided data set in \xs{sect:data-description}.

\begin{figure}[htpb]%
	\centering
	\includegraphics[width=.7\columnwidth]{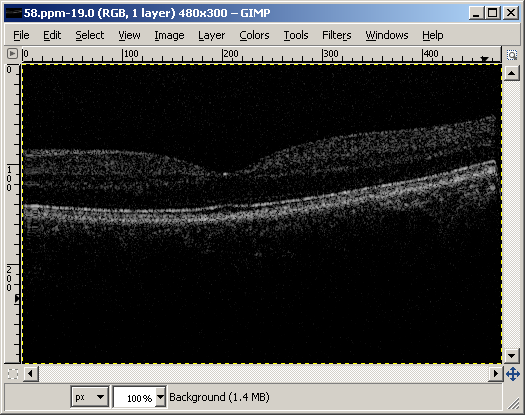}%
	\caption{Test of Loading Provided Data in PPM To Match Provided}%
	\label{fig:slice58-ppm-reproduced-simple}%
\end{figure}

\begin{figure}[htpb]%
	\centering
	\includegraphics[width=.7\columnwidth]{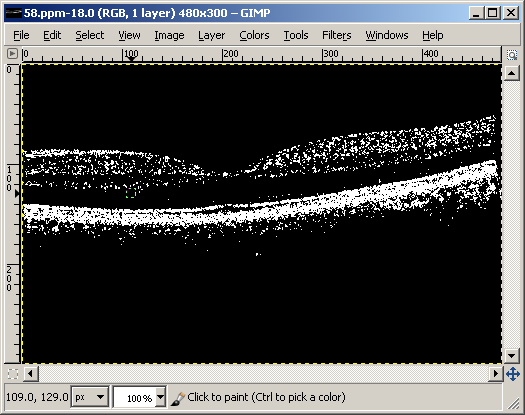}%
	\caption{Test of Rudimentary Processing via Threshold}%
	\label{fig:slice58-bw-denoised-simple}%
\end{figure}

\subsection{APIs}

\subsubsection{Internal}

The internal API at the moment centers around the detector/analyzer/preprocessor
frameworks and is invoked by \api{OCTMARFApp}. Most of them take the \api{Volume}
data structure and do something with it in either 2D or 3D or both. The API provides for both working on
the actual \api{Volume} instance or its copy if desired to preserve the original.

\begin{itemize}
	\item \api{IDetector} -- specifies the API all boundary detectors should implement
	\item \api{Detector} -- generic abstract class that concrete classes may inherit from for convenience
	\item \api{FabritiusILMDetector} -- a specific detector being implemented
	\item \api{ZhanSuenTransform} -- an example concrete detector and a processor that relies on the implementation of the \cite{zhang-suen-transform-1984}
	\item \api{IPreprocessor} -- defines API for all preprocessors (e.g. binarizers, denoisers, or filters or whatever) to use
	\item \api{Preprocessor} -- a generic class implementing rudimentary API for convenience
	\item \api{Binarizer} -- a concrete simple preprocessor binarizing images with a set threshold
	\item \api{IProcessor} -- API for all kinds of analyzers
	\item \api{Configuration} -- flags and settings for experiment automation
	\item ...
\end{itemize}

\subsubsection{External}

External API has to do with libraries, frameworks, used or planned to be
used for evaluation, processing, or connectivity with other platforms
and languages.

\begin{itemize}
	\item MARF \cite{marf}
	\item JAI
	\item JNI/JNA \cite{jni}
	\item GIPSY \cite{gipsy}
\end{itemize}

\section{Experiments}
\label{sect:experiments}

Some experiments are in place to test the data processing, the framework's API
operation, and some actual algorithms.
That includes loading of the \texttt{.txt} data and its subsequent dump back as an image work
(see \xf{fig:slice58-ppm-reproduced-simple})
as well as testing the data structure by rudimentary processing of the loaded
data's pixels by simply discarding values below certain intensity threshold and
amplifying the remaining ones to the max as a very basic test
(see \xf{fig:slice58-bw-denoised-simple}) to ensure the given data understanding
is correct via \api{Binarizer}.

More interesting preliminary results include global 2D (2D at the moment, per
slice, but will expand the window to 3D space next) processing that includes
in one algorithm some subalgorithms that do preprocessing, skeletonizing,
or filling/thinning, and active-contouring of the images from the Zhang-Suen
transform. There are many of the options and configurable parameters to try,
but already now the outlines of the ILM, IS/OS, RPE can be seen in a very
draft debug mode in \xf{fig:slice58-zsuen-countour}, \xf{fig:slice90-zsuen-countour},
\xf{fig:zsuen-skel-58}, \xf{fig:zsuen-skel-90}, \xf{fig:zsuen-fts-3-58},
and \xf{fig:zsuen-fts-3-90}.

Framework accepts \api{Configuration} object where multiple configurations
with or without preprocessing can be tried on the whole volume.

These can be further refined with parameter tweaking as well as other
algorithms (e.g. Canny edge detection, graph search, etc.).

Run-times for the current experiments varied from 29 seconds to 5
minutes depending on the configuration for the whole 3D volume.

\begin{figure}[htpb]%
	\centering
	\includegraphics[width=.7\columnwidth]{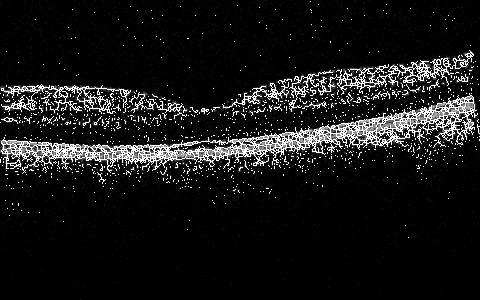}%
	\caption{2D Zhang-Suen Skeletonizing near Fovea}%
	\label{fig:zsuen-skel-58}%
\end{figure}

\begin{figure}[htpb]%
	\centering
	\includegraphics[width=.7\columnwidth]{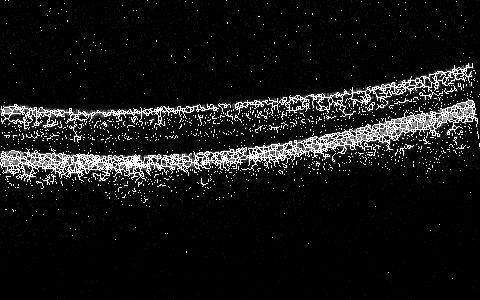}%
	\caption{2D Zhang-Suen Skeletonizing of Slice 90}%
	\label{fig:zsuen-skel-90}%
\end{figure}

\begin{figure}[htpb]%
	\centering
	\includegraphics[width=.7\columnwidth]{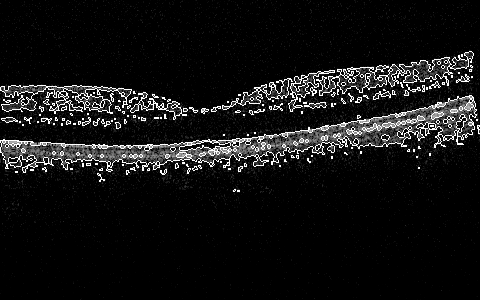}%
	\caption{2D Zhang-Suen Filling-Thinning-Contouring near Fovea}%
	\label{fig:slice58-zsuen-countour}%
\end{figure}

\begin{figure}[htpb]%
	\centering
	\includegraphics[width=.7\columnwidth]{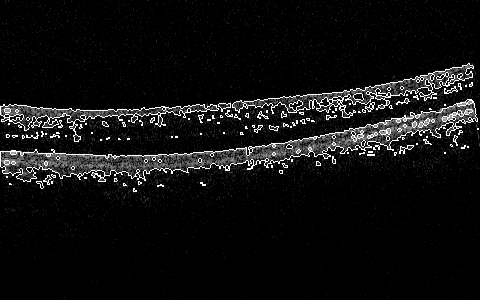}%
	\caption{2D Zhang-Suen Filling-Thinning-Contouring of Slice 90}%
	\label{fig:slice90-zsuen-countour}%
\end{figure}

\begin{figure}[htpb]%
	\centering
	\includegraphics[width=.7\columnwidth]{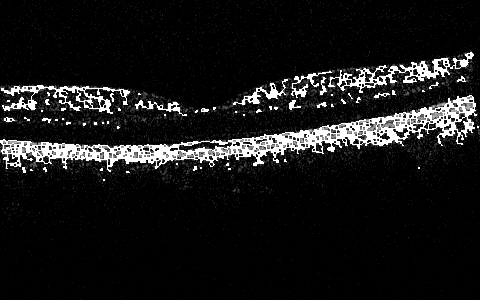}%
	\caption{2D Zhang-Suen Filling-Thinning-Skeletonizing near Fovea}%
	\label{fig:zsuen-fts-3-58}%
\end{figure}

\begin{figure}[htpb]%
	\centering
	\includegraphics[width=.7\columnwidth]{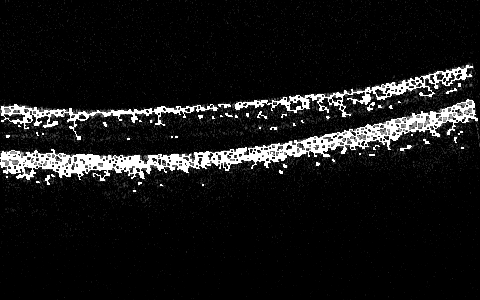}%
	\caption{2D Zhang-Suen Filling-Thinning-Skeletonizing of Slice 90}%
	\label{fig:zsuen-fts-3-90}%
\end{figure}

\section*{Acknowledgments}

Serguei Mokhov is supported in part by the CCSEP scholarship and the Faculty of Engineering
and Computer Science (ENCS), Concordia University, Montreal, Canada.
Yankui Sun's research is supported by the National Natural Science Foundation of China 
no. 60971006.

\addcontentsline{toc}{section}{References}
\nocite{cryptolysis-inse6110-05}
\bibliography{oct-review}

\appendix

\section{Proof-of-Concept Code Excerpts}
\label{appdx:code-excerpts}

\subsection{\api{AScanLoader}}
\scriptsize
\begin{Verbatim}
package marf.apps.oct.OCTMARF.Storage;

import java.io.BufferedReader;
import java.io.File;
import java.io.FileReader;
import java.io.IOException;

import marf.apps.oct.OCTMARF.framework.processors.ZhangSuenTransform;
import marf.util.Arrays;

/**
 * Loads and parses A-scans from .txt files into internal data structures.
 * 
 * @author Serguei Mokhov
 * @since 0.0.1, October 30, 2011
 * @version $Id: AScanLoader.java,v 1.8 2012/01/02 08:06:32 mokhov Exp $
 */
public class AScanLoader
{
	/**
	 * 3D image volume storage.
	 */
	protected Volume oVolume = new Volume();
	
	/**
	 * Current slice to process.
	 * Usually during loading.
	 */
	protected int iScanSlice = 0;

	/**
	 * Empty loader.
	 */
	public AScanLoader()
	{
	}

	/**
	 * Should the scan fail to load.
	 * @param pstrSampleFilename
	 * @throws IOException
	 * @see {@link #load(String)}
	 */
	public AScanLoader(String pstrSampleFilename)
	throws IOException
	{
		load(pstrSampleFilename);
	}
	
	/**
	 * Loads the current specified slice.
	 * @param pstrSampleFilename
	 * @throws IOException
	 */
	public void load(String pstrSampleFilename)
	throws IOException
	{
		BufferedReader oBFR = new BufferedReader(new FileReader(pstrSampleFilename));
		String strDataLine = oBFR.readLine();
		
		int iAScanLine = 0;
		
		while(strDataLine != null)
		{
			//System.out.println("Data line: [" + strDataLine + "]");
			
			String[] astrPixels = strDataLine.split(" ");
			//System.out.println("Data line element count: [" + astrPixels.length + "]");

			//this.astr3DScan[this.iScanSlice][iAScanLine] = astrPixels;
			//Arrays.copy(this.as3DScan[this.iScanSlice][iAScanLine], astrPixels);
			//Arrays.copy(this.ad3DScan[this.iScanSlice][iAScanLine], this.as3DScan[this.iScanSlice][iAScanLine]);

			this.oVolume.astr3DScan[this.iScanSlice][iAScanLine] = astrPixels;
			Arrays.copy(this.oVolume.as3DScan[this.iScanSlice][iAScanLine], astrPixels);
			Arrays.copy(this.oVolume.ad3DScan[this.iScanSlice][iAScanLine], this.oVolume.as3DScan[this.iScanSlice][iAScanLine]);
			
			iAScanLine++;
			strDataLine = oBFR.readLine();
			//strDataLine = null;
		}
		
		oBFR.close();

		// Move to the next slice for (potential) loading
		this.iScanSlice++;
	}
	
	/**
	 * @return
	 */
	public short[][][] getShort3DScan()
	{
		return this.oVolume.as3DScan;
	}

	/**
	 * @return
	 */
	public double[][][] getDouble3DScan()
	{
		return this.oVolume.ad3DScan;
	}

	/**
	 * @return
	 */
	public String[][][] getString3DScan()
	{
		return this.oVolume.astr3DScan;
	}

	public Volume getVolume()
	{
		return this.oVolume;
	}
	
	public int sizeInSlices()
	{
		return this.iScanSlice + 1;
	}
}

// EOF
\end{Verbatim}
\normalsize

\subsection{\api{Volume}}
\scriptsize
\begin{Verbatim}
package marf.apps.oct.OCTMARF.Storage;

import java.io.Serializable;

/**
 * TODO: document.
 *
 *
 * $Id: Volume.java,v 1.1 2011/12/18 07:17:34 mokhov Exp $
 *
 * @author serguei
 * @since
 */
public class Volume
implements Serializable, Cloneable
{
	/**
	 * 
	 */
	private static final long serialVersionUID = -6995648761632634728L;

	/**
	 * A-scan's line.
	 */
	public static final int DEFAULT_SCAN_HEIGHT = 300;
	
	/**
	 * Normal scan dimension. 
	 */
	public static final int DEFAULT_SCAN_WIDTH = 480;
	
	/**
	 * Number of scan slices.
	 */
	public static final int DEFAULT_SCAN_DEPTH = 100;

	protected short[][][] as3DScan = new short[DEFAULT_SCAN_DEPTH][DEFAULT_SCAN_WIDTH][DEFAULT_SCAN_HEIGHT];
	protected double[][][] ad3DScan = new double[DEFAULT_SCAN_DEPTH][DEFAULT_SCAN_WIDTH][DEFAULT_SCAN_HEIGHT];
	protected String[][][] astr3DScan = new String[DEFAULT_SCAN_DEPTH][DEFAULT_SCAN_WIDTH][DEFAULT_SCAN_HEIGHT];


	private boolean bEmpty = true;
	
	public Volume()
	{
		super();
	}

	/**
	 * @return the as3DScan
	 */
	public short[][][] getShort3DScan()
	{
		return as3DScan;
	}

	/**
	 * @param as3dScan the as3DScan to set
	 */
	public void setShort3DScan(short[][][] as3dScan) {
		as3DScan = as3dScan;
	}

	/**
	 * @return the ad3DScan
	 */
	public double[][][] getDouble3DScan() {
		return ad3DScan;
	}

	/**
	 * @param ad3dScan the ad3DScan to set
	 */
	public void setDouble3DScan(double[][][] ad3dScan) {
		ad3DScan = ad3dScan;
	}

	/**
	 * @return the astr3DScan
	 */
	public String[][][] getString3DScan() {
		return astr3DScan;
	}

	/**
	 * @param astr3dScan the astr3DScan to set
	 */
	public void setString3DScan(String[][][] astr3dScan) {
		astr3DScan = astr3dScan;
	}

	public boolean isEmpty()
	{
		return this.bEmpty;
	}

	/* (non-Javadoc)
	 * @see java.lang.Object#clone()
	 */
	@Override
	public Object clone()
	//throws CloneNotSupportedException
	{
		Volume oCopy = null;
		
		try
		{
			oCopy = (Volume)super.clone();
		
			if(this.ad3DScan != null)
			{
				oCopy.ad3DScan = this.ad3DScan.clone();
			}
			
			if(this.as3DScan != null)
			{
				oCopy.as3DScan = this.as3DScan.clone();
			}
			
			if(this.astr3DScan != null)
			{
				oCopy.astr3DScan = this.astr3DScan.clone();
			}
		}
		catch(CloneNotSupportedException e)
		{
			System.err.println("Failed to clone volume.");
			e.printStackTrace(System.err);
		}
		
		return oCopy;
	}

	/* (non-Javadoc)
	 * @see java.lang.Object#finalize()
	 */
	@Override
	protected void finalize()
	throws Throwable
	{
		this.ad3DScan = null;
		this.as3DScan = null;
		this.astr3DScan = null;
		super.finalize();
	}

	/* (non-Javadoc)
	 * @see java.lang.Object#toString()
	 */
	@Override
	public String toString()
	{
		return "}: empty: " + this.bEmpty;
	}
}

// EOF
\end{Verbatim}
\normalsize

\subsection{\api{OCTMARFApp}}
\scriptsize
\begin{Verbatim}
package marf.apps.oct.OCTMARF;

import java.io.File;
import java.util.ArrayList;

import marf.MARF;
import marf.Classification.Distance.HammingDistance;
import marf.Storage.ModuleParams;
import marf.Storage.TrainingSet;
import marf.apps.oct.OCTMARF.Storage.AScanLoader;
import marf.apps.oct.OCTMARF.Storage.AnyToWAVLoader;
import marf.apps.oct.OCTMARF.Storage.PPMDumper;
import marf.apps.oct.OCTMARF.Storage.RawSampleLoader;
import marf.apps.oct.OCTMARF.framework.preprocessing.Binarizer;
import marf.apps.oct.OCTMARF.framework.processors.ZhangSuenTransform;
import marf.util.Debug;
import marf.util.MARFException;
import marf.util.NotImplementedException;
import marf.util.OptionProcessor;


/**
 * <p>
 * A an OCT image processing testbed platform for various algorithms. 
 * </p>
 * 
 * @author Serguei Mokhov
 * 
 * @version $Id: OCTMARFApp.java,v 1.6 2012/01/02 08:06:32 mokhov Exp $
 * @since October 2011
 */
public class OCTMARFApp
{
	/*
	 * ---------------- Apps. Versioning ----------------
	 */

	/**
	 * Current major version of the application.
	 */
	public static final int MAJOR_VERSION = 0;

	/**
	 * Current minor version of the application.
	 */
	public static final int MINOR_VERSION = 0;

	/**
	 * Current revision of the application.
	 */
	public static final int REVISION = 1;

	/*
	 * ----------------------------------
	 * Defaults
	 * ----------------------------------
	 */

	/**
	 * Default file name extension for sample files; case-insensitive.
	 */
	public static final String DEFAULT_SAMPLE_FILE_EXTENSION = "txt";

	/*
	 * ----------------------------------
	 * Major and Misc Options Enumeration
	 * ----------------------------------
	 */

	/**
	 * Numeric equivalent of the option <code>--train</code>.
	 */
	public static final int OPT_TRAIN = 0;

	/**
	 * Numeric equivalent of the option <code>--ident</code>.
	 */
	public static final int OPT_IDENT = 1;

	/**
	 * Numeric equivalent of the option <code>--stats</code>.
	 */
	public static final int OPT_STATS = 2;

	/**
	 * Numeric equivalent of the option <code>--reset</code>.
	 */
	public static final int OPT_RESET = 3;

	/**
	 * Numeric equivalent of the option <code>--version</code>.
	 */
	public static final int OPT_VERSION = 4;

	/**
	 * Numeric equivalent of the option <code>--help</code>.
	 */
	public static final int OPT_HELP_LONG = 5;

	/**
	 * Numeric equivalent of the option <code>-h</code>.
	 */
	public static final int OPT_HELP_SHORT = 6;

	/**
	 * Numeric equivalent of the option <code>-debug</code>.
	 */
	public static final int OPT_DEBUG = 7;

	/**
	 * Numeric equivalent of the option <code>-graph</code>.
	 */
	public static final int OPT_GRAPH = 8;

	/**
	 * Numeric equivalent of the option <code>-spectrogram</code>.
	 */
	public static final int OPT_SPECTROGRAM = 9;

	/**
	 * Numeric equivalent of the option <code>&lt;writer ID&gt;</code>.
	 */
	public static final int OPT_EXPECTED_ID = 10;

	/**
	 * Numeric equivalent of the option <code>--batch-ident</code>.
	 */
	public static final int OPT_BATCH_IDENT = 11;

	/**
	 * Numeric equivalent of the option <code>--single-train</code>.
	 */
	public static final int OPT_SINGLE_TRAIN = 12;

	/**
	 * Numeric equivalent of the option
	 * <code>&lt;sample-file-or-directory-name&gt;</code>.
	 */
	public static final int OPT_DIR_OR_FILE = 13;

	/**
	 * Numeric equivalent of the option <code>--best-score</code>.
	 */
	public static final int OPT_BEST_SCORE = 14;

	/**
	 * Numeric equivalent of the option <code>--gui</code> to have a GUI.
	 */
	public static final int OPT_GUI = 15;

	/**
	 * Numeric equivalent of the option <code>-noise</code> to enable noise
	 * removal.
	 */
	public static final int OPT_NOISE_REMOVAL = 16;

	/**
	 * Numeric equivalent of the option <code>-silence</code> to enable silence
	 * removal.
	 */
	public static final int OPT_SILENCE_REMOVAL = 17;

	/**
	 * Numeric equivalent of the option <code>--show-training-set</code> to
	 * display text representation of a training set file.
	 */
	public static final int OPT_SHOW_TRAINING_SET_DATA = 18;

	public static final int OPT_NO_PREPREP = 20;

	public static final int OPT_SILENCE_NOISE_REMOVAL = 27;

	public static final int OPT_TRAINING_MODEL_MEAN_CLUSTER = 29;
	public static final int OPT_TRAINING_MODEL_MEDIAN_CLUSTER = 30;
	public static final int OPT_TRAINING_MODEL_FEATURE_SET = 31;

	/*
	 * Some defaults
	 */
	
	public static final Integer DEFAULT_SEVERITY = 2;
	public static final Double DEFAULT_PROBABILITY = 0.0;

	public static final Double DISTANCE_THRESHOLD = 0.1;
	public static final Double SIMILARITY_THRESHOLD = 0.001;
	public static final Double PROBABILITY_THRESHOLD = 0.9;

	/*
	 * ---------------------
	 * State Data Structures
	 * ---------------------
	 */

	/**
	 * Instance of the option processing utility.
	 */
	protected static OptionProcessor soGetOpt = new OptionProcessor();

	/*
	 * ----------------- Static State Init -----------------
	 */

	static
	{
		// Main options
		soGetOpt.addValidOption(OPT_TRAIN, "--train");
		soGetOpt.addValidOption(OPT_SINGLE_TRAIN, "--single-train");
		soGetOpt.addValidOption(OPT_IDENT, "--ident");
		soGetOpt.addValidOption(OPT_BATCH_IDENT, "--batch-ident");
		soGetOpt.addValidOption(OPT_STATS, "--stats");
		soGetOpt.addValidOption(OPT_BEST_SCORE, "--best-score");
		soGetOpt.addValidOption(OPT_SHOW_TRAINING_SET_DATA, "--show-training-set");

		soGetOpt.addValidOption(OPT_VERSION, "--version");
		soGetOpt.addValidOption(OPT_RESET, "--reset");
		soGetOpt.addValidOption(OPT_HELP_LONG, "--help");
		soGetOpt.addValidOption(OPT_HELP_SHORT, "-h");
		soGetOpt.addValidOption(OPT_DEBUG, "--debug");

		// Sample Loading
		soGetOpt.addValidOption(MARF.TEXT, "-text");

		// Preprocessing
		soGetOpt.addValidOption(OPT_NOISE_REMOVAL, "-noise");
		soGetOpt.addValidOption(OPT_SILENCE_REMOVAL, "-silence");
		soGetOpt.addValidOption(OPT_SILENCE_NOISE_REMOVAL, "-silence-noise");

		soGetOpt.addValidOption(MARF.DUMMY, "-norm");
		soGetOpt.addValidOption(MARF.HIGH_FREQUENCY_BOOST_FFT_FILTER, "-boost");
		soGetOpt.addValidOption(MARF.HIGH_PASS_FFT_FILTER, "-high");
		soGetOpt.addValidOption(MARF.LOW_PASS_FFT_FILTER, "-low");
		soGetOpt.addValidOption(MARF.BANDPASS_FFT_FILTER, "-band");
		soGetOpt.addValidOption(MARF.BAND_STOP_FFT_FILTER, "-bandstop");
		soGetOpt.addValidOption(MARF.HIGH_PASS_BOOST_FILTER, "-highpassboost");
		soGetOpt.addValidOption(MARF.RAW, "-raw");
		soGetOpt.addValidOption(MARF.ENDPOINT, "-endp");
		soGetOpt.addValidOption(MARF.HIGH_PASS_CFE_FILTER, "-highcfe");
		soGetOpt.addValidOption(MARF.LOW_PASS_CFE_FILTER, "-lowcfe");
		soGetOpt.addValidOption(MARF.BAND_PASS_CFE_FILTER, "-bandcfe");
		soGetOpt.addValidOption(MARF.BAND_STOP_CFE_FILTER, "-bandstopcfe");

		// Wavelets
		soGetOpt.addValidOption(MARF.SEPARABLE_DWT_FILTER, "-sdwt");
		soGetOpt.addValidOption(MARF.DUAL_DTREE_DWT_FILTER, "-dualtree");
		soGetOpt.addValidOption(MARF.DYADIC_DWT_FILTER, "-dyadic");
		
		// Feature extraction
		soGetOpt.addValidOption(MARF.FFT, "-fft");
		soGetOpt.addValidOption(MARF.LPC, "-lpc");
		soGetOpt.addValidOption(MARF.RANDOM_FEATURE_EXTRACTION, "-randfe");
		soGetOpt.addValidOption(MARF.MIN_MAX_AMPLITUDES, "-minmax");
		soGetOpt.addValidOption(MARF.FEATURE_EXTRACTION_AGGREGATOR, "-aggr");
		soGetOpt.addValidOption(MARF.F0, "-f0");
		soGetOpt.addValidOption(MARF.SEGMENTATION, "-segm");
		soGetOpt.addValidOption(MARF.CEPSTRAL, "-cepstral");

		// Classification
		soGetOpt.addValidOption(MARF.NEURAL_NETWORK, "-nn");
		soGetOpt.addValidOption(MARF.EUCLIDEAN_DISTANCE, "-eucl");
		soGetOpt.addValidOption(MARF.CHEBYSHEV_DISTANCE, "-cheb");
		soGetOpt.addValidOption(MARF.MINKOWSKI_DISTANCE, "-mink");
		soGetOpt.addValidOption(MARF.MAHALANOBIS_DISTANCE, "-mah");
		soGetOpt.addValidOption(MARF.RANDOM_CLASSIFICATION, "-randcl");
		soGetOpt.addValidOption(MARF.DIFF_DISTANCE, "-diff");
		soGetOpt.addValidOption(MARF.ZIPFS_LAW, "-zipf");
		soGetOpt.addValidOption(MARF.MARKOV, "-markov");
		soGetOpt.addValidOption(MARF.HAMMING_DISTANCE, "-hamming");
		soGetOpt.addValidOption(MARF.COSINE_SIMILARITY_MEASURE, "-cos");

		// Graphical
		soGetOpt.addValidOption(OPT_SPECTROGRAM, "-spectrogram");
		soGetOpt.addValidOption(OPT_GRAPH, "-graph");
		soGetOpt.addValidOption(OPT_GUI, "--gui");

		// NLP
		soGetOpt.addValidOption(MARF.EStatisticalEstimators.MLE, "-mle");
		soGetOpt.addValidOption(MARF.EStatisticalEstimators.ADD_ONE, "-add-one");
		soGetOpt.addValidOption(MARF.EStatisticalEstimators.ADD_DELTA, "-add-delta");
		soGetOpt.addValidOption(MARF.EStatisticalEstimators.ELE, "-ele");
		soGetOpt.addValidOption(MARF.EStatisticalEstimators.WITTEN_BELL, "-witten-bell");
		soGetOpt.addValidOption(MARF.EStatisticalEstimators.GOOD_TURING, "-good-turing");
		soGetOpt.addValidOption(MARF.EStatisticalEstimators.SLI, "-sli");
		soGetOpt.addValidOption(MARF.EStatisticalEstimators.GLI, "-gli");
		soGetOpt.addValidOption(MARF.EStatisticalEstimators.KATZ_BACKOFF, "-backoff");

		soGetOpt.addValidOption(MARF.NLP.STEMMING, "-stemming");

		soGetOpt.addValidOption(MARF.NLP.CHARACTER_MODE, "-char");
		soGetOpt.addValidOption(MARF.NLP.WORD_MODE, "-word");

		soGetOpt.addValidOption(MARF.NLP.CASE_SENSITIVE, "-case");
		soGetOpt.addValidOption(MARF.NLP.PARSE_NUMBERS, "-num");
		soGetOpt.addValidOption(MARF.NLP.PARSE_QUOTED_STRINGS, "-quote");
		soGetOpt.addValidOption(MARF.NLP.PARSE_ENDS_OF_SENTENCE, "-eos");
		soGetOpt.addValidOption(MARF.NLP.RAW_ZIPFS_LAW_DUMP, "-nolog");

		soGetOpt.addValidOption(MARF.ENgramModels.UNIGRAM, "-unigram");
		soGetOpt.addValidOption(MARF.ENgramModels.BIGRAM, "-bigram");
		soGetOpt.addValidOption(MARF.ENgramModels.TRIGRAM, "-trigram");

		soGetOpt.addValidOption(MARF.NLP.TRAIN, "--train-nlp");
		soGetOpt.addValidOption(MARF.NLP.CLASSIFY, "--ident-nlp");

		soGetOpt.addValidOption(MARF.NLP.ZIPFS_LAW_CHEAT, "--cheat");

		soGetOpt.addValidOption(MARF.NLP.INTERACTIVE, "-interactive");

		soGetOpt.addValidOption(OPT_NO_PREPREP, "-nopreprep");
	}

	public static class Configuration
	{
		String strFilePrefix = "";
		boolean bBinarize = false;
		boolean bZSSkeletonizing = false;
		boolean bZSFilling = false;
		boolean bZSThinning = false;
		boolean bZSCountouring = false;
		
		int iRepeat = 1;
		
		public Configuration()
		{
			
		}
		
		public String toString()
		{
			return strFilePrefix + ":" + bBinarize + ":" + bZSSkeletonizing + ":" + bZSFilling + ":" + bZSThinning + ":" + bZSCountouring;
		}
	}
	
	/**
	 * Main body.
	 * 
	 * @param argv command-line arguments
	 */
	public static final void main(String[] argv)
	{
		try
		{
			// Since some new API is always introduced...
			validateVersions();

			// Parse extra arguments
			int iValidOptions = soGetOpt.parse(argv);

			// Turn on command-line supplied debugging
			Debug.enableDebug(soGetOpt.isActiveOption(OPT_DEBUG));
			Debug.debug("Option set parsed: " + soGetOpt);

			if(iValidOptions == 0)
			{
				throw new Exception("No valid options found: " + soGetOpt);
			}

			switch(soGetOpt.getInvalidOptions().size())
			{
				case 0:
				{
					break;
				}

				/*
				 * In the case of a single invalid option always assume it
				 * is either a filename or a directory name for the --ident
				 * or --train options.
				 */
				case 1:
				{
					soGetOpt.addActiveOption(OPT_DIR_OR_FILE, soGetOpt.getInvalidOptions().firstElement().toString());
					soGetOpt.getInvalidOptions().clear();
					break;
				}

				default:
				{
					throw new Exception("Unrecognized options found: " + soGetOpt.getInvalidOptions());
				}
			}

			setDefaultConfig();
			setCustomConfig();

			// Set misc configuration
			MARF.setDumpSpectrogram(soGetOpt.isActiveOption(OPT_SPECTROGRAM));
			MARF.setDumpWaveGraph(soGetOpt.isActiveOption(OPT_GRAPH));

			Debug.debug("Option set: " + soGetOpt);

			if(soGetOpt.isActiveOption(MARF.TEXT))
			{
				MARF.setSampleFormat(MARF.TEXT);
				MARF.setSampleLoaderPluginClass(AnyToWAVLoader.class.getName());
				System.out.println("Using text loader");
			}
			else if(soGetOpt.isActiveOption(MARF.TEXT + 2))
			{
				MARF.setSampleFormat(MARF.CUSTOM);
				// XXX:
				MARF.setSampleLoaderPluginClass(RawSampleLoader.class.getName());
			}

			int iMainOption = soGetOpt.getOption(argv[0]);

			switch(iMainOption)
			{
				/*
				 * --------------
				 * Identification
				 * --------------
				 */

				// Single case
				case OPT_IDENT:
				case OPT_BATCH_IDENT:
				case MARF.NLP.CLASSIFY:
				{
					ArrayList<Configuration> oConfigs = new ArrayList<Configuration>();
					
					Configuration oConfig = new Configuration();
					oConfig.bBinarize = false;
					oConfig.bZSCountouring = false;
					oConfig.bZSFilling = false;
					oConfig.bZSThinning = false;
					oConfig.bZSSkeletonizing = true;
					oConfig.strFilePrefix = "zsuen-skel";
					oConfig.iRepeat = 1;
					oConfigs.add(oConfig);

					oConfig = new Configuration();
					oConfig.bBinarize = true;
					oConfig.bZSCountouring = false;
					oConfig.bZSFilling = false;
					oConfig.bZSThinning = false;
					oConfig.bZSSkeletonizing = false;
					oConfig.strFilePrefix = "simple-bin";
					oConfig.iRepeat = 1;
					oConfigs.add(oConfig);
					
					oConfig = new Configuration();
					oConfig.bBinarize = false;
					oConfig.bZSCountouring = true;
					oConfig.bZSFilling = true;
					oConfig.bZSThinning = true;
					oConfig.bZSSkeletonizing = false;
					oConfig.strFilePrefix = "zsuen-ftc";
					oConfig.iRepeat = 1;
					oConfigs.add(oConfig);

					oConfig = new Configuration();
					oConfig.bBinarize = false;
					oConfig.bZSCountouring = false;
					oConfig.bZSFilling = true;
					oConfig.bZSThinning = true;
					oConfig.bZSSkeletonizing = true;
					oConfig.strFilePrefix = "zsuen-fts-3";
					oConfig.iRepeat = 3;
					oConfigs.add(oConfig);

					/*
					oConfig = new Configuration();
					oConfig.bBinarize = false;
					oConfig.bZSCountouring = false;
					oConfig.bZSFilling = true;
					oConfig.bZSThinning = false;
					oConfig.bZSSkeletonizing = true;
					oConfig.strFilePrefix = "zsuen-fs-3";
					oConfig.iRepeat = 3;
					oConfigs.add(oConfig);
					*/

					for(Configuration oRunConfig: oConfigs)
					{
						ident(oRunConfig);
					}
					
					break;
				}

				/*
				 * --------
				 * Training
				 * --------
				 */

				// Add a single sample to the training set
				case OPT_SINGLE_TRAIN:
				{
					throw new NotImplementedException();
				}

				// Train on a directory of files
				case OPT_TRAIN:
				case MARF.NLP.TRAIN:
				{
					throw new NotImplementedException();
				}

				/*
				 * -------------------------
				 * Training Set Data Display
				 * -------------------------
				 */
				case OPT_SHOW_TRAINING_SET_DATA:
				{
					System.out.println
					(
						marf.Classification.Classification.loadTrainingSet
						(
							marf.Storage.IStorageManager.DUMP_GZIP_BINARY,
							soGetOpt.getOption(OPT_DIR_OR_FILE)
						)
					);

					break;
				}

				/*
				 * -----
				 * Stats
				 * -----
				 */
				case OPT_STATS:
				{
					throw new NotImplementedException();
				}

				/*
				 * Best Result with Stats
				 */
				case OPT_BEST_SCORE:
				{
					throw new NotImplementedException();
				}

				/*
				 * Reset Stats
				 */
				case OPT_RESET:
				{
					throw new NotImplementedException();
				}

				/*
				 * Versioning
				 */
				case OPT_VERSION:
				{
					System.out.println("OCTMARF Application, v." + getVersion());
					System.out.println("Using MARF, v." + MARF.getVersion());
					validateVersions();
					break;
				}

				/*
				 * Help
				 */
				case OPT_HELP_SHORT:
				case OPT_HELP_LONG:
				{
					usage();
					break;
				}

				/*
				 * Invalid major option
				 */
				default:
				{
					throw new Exception("Unrecognized option: " + argv[0]);
				}
			} // major option switch
		} // try

		/*
		 * No arguments have been specified
		 */
		catch(ArrayIndexOutOfBoundsException e)
		{
			System.err.println("No arguments have been specified.");
			System.err.println(e.getMessage());
			e.printStackTrace(System.err);

			usage();
		}

		/*
		 * MARF-specific errors
		 */
		catch(MARFException e)
		{
			System.err.println(e.getMessage());
			e.printStackTrace(System.err);
		}

		/*
		 * Invalid option and/or option argument
		 */
		catch(Exception e)
		{
			System.err.println(e.getMessage());
			e.printStackTrace(System.err);
			usage();
		}

		/*
		 * Regardless whatever happens, close the db connection.
		 */
		finally
		{
			try
			{
				Debug.debug("Closing DB connection...");
			}
			catch(Exception e)
			{
				Debug.debug("Closing DB connection failed: " + e.getMessage());
				e.printStackTrace(System.err);
				System.exit(-1);
			}
		}
	}

	/**
	 * Computes duration and outputs it as a string
	 * @param plStartTime
	 * @param plEndTime
	 * @return String-formatted duration
	 */
	private static String getDuration(long plStartTime, long plEndTime)
	{
		// XXX: constify
		int iOneSecInMs  = 1000;
		int iOneMinInMs  = 60 * iOneSecInMs;
		int iOneHourInMs = 60 * iOneMinInMs;
		int iOneDayInMs  = 24 * iOneHourInMs;

		long lDuration = plEndTime - plStartTime;

		long lDays = lDuration / iOneDayInMs;
		long lModTimeLeft = lDuration % iOneDayInMs;

		long lHours = lModTimeLeft / iOneHourInMs;
		lModTimeLeft %= iOneHourInMs;

		long lMinutes = lModTimeLeft / iOneMinInMs;
		lModTimeLeft %= iOneMinInMs;

		long lSeconds = lModTimeLeft / iOneSecInMs;
		lModTimeLeft %= iOneSecInMs;

		long lMilliSeconds = lModTimeLeft;

		StringBuffer oDuration = new StringBuffer()
			.append(lDays + "d:" + lHours + "h:" + lMinutes + "m:" + lSeconds +"s:" + lMilliSeconds +"ms:" + lDuration + "ms")
			;

		return oDuration.toString();
	}
	
	/**
	 * Generic ident routine.
	 * @throws Exception
	 */
	public static final void ident(Configuration poConfig)
	throws Exception
	{
		long lTimeStart = System.currentTimeMillis();
		{
			AScanLoader oLoader = new AScanLoader();
			
			// Process all .txt files in the directory
			File[] aoSampleFiles = new File("resources/data").listFiles();

			Binarizer oBinarizer = new Binarizer();						
			ZhangSuenTransform oTransform = new ZhangSuenTransform();

			for(int i = 0; i < aoSampleFiles.length; i++)
			{
				String strFilename = aoSampleFiles[i].getPath();

				if(aoSampleFiles[i].isFile() && strFilename.toLowerCase().endsWith(DEFAULT_SAMPLE_FILE_EXTENSION))
				{
					long lSliceTimeStart = System.currentTimeMillis();
					
					// Load them one by one
					System.out.println("Loading " + strFilename + " into slice " + oLoader.sizeInSlices() + " of 3D volume.");
					oLoader.load(strFilename);
					
					// Trial processing
					{
						PPMDumper oDumper = new PPMDumper(poConfig.strFilePrefix + "." + i);
						
						if(poConfig.bBinarize == true)
						{
							oBinarizer.preprocess2D(oLoader.getVolume(), i);
						}
						
						if
						(
							poConfig.bZSCountouring == true
							||
							poConfig.bZSFilling == true
							||
							poConfig.bZSSkeletonizing == true
							||
							poConfig.bZSThinning == true
						)
						{
							oTransform.setCountouring(poConfig.bZSCountouring);
							oTransform.setFilling(poConfig.bZSFilling);
							oTransform.setSkeletonizing(poConfig.bZSSkeletonizing);
							oTransform.setThinning(poConfig.bZSThinning);
							
							for(int r = 0; r < poConfig.iRepeat; r++)
							{
								oTransform.detect2D(oLoader.getVolume(), i);
							}
						}
						
						System.out.print("Dumping " + oDumper.getFilename() + "... ");
						for(double[] adScanLine: oLoader.getDouble3DScan()[i])
						{
							oDumper.addToData(adScanLine);
						}
					
						// Save as a PPM
						oDumper.dump();
						System.out.println("Done.");
					} // end trial processing 

					long lSliceTimeFinish = System.currentTimeMillis();
					System.out.println
					(
						"Slice processing and saving time: " +
						getDuration(lSliceTimeStart, lSliceTimeFinish)
					);
				} // if
			}
		}
		long lTimeFinish = System.currentTimeMillis();
		System.out.println
		(
			"Total time volume took: " + getDuration(lTimeStart, lTimeFinish)
			+ " for configuration: " + poConfig
		);
	}

	/**
	 * Displays application's usage information and exits.
	 */
	private static final void usage()
	{
		System.out.println
		(
			"Usage:\n"
			+ "  java OCTMARFApp      --train <samples-dir> [options]        -- train mode\n"
			+ "                       --single-train <sample> [options]      -- add a single sample to the training set\n"
			+ "                       --ident <sample> [options]             -- identification mode\n"
			+ "                       --batch-ident <samples-dir> [options]  -- batch identification mode\n"
			+ "                       --train-nlp [ --debug ] [ OPTIONS ] <language> <corpus-file>\n"
			+ "                       --ident-nlp [ --debug ] [ OPTIONS ] foo <bar|corpus-file>\n\n"
			+ "                       --debug                                -- include verbose debug output\n"
			+ "                       --gui                                  -- use GUI as a user interface\n"
			+ "                       --stats=[per-config|per-writer|both]   -- display stats (default is per-config)\n"
			+ "                       --best-score                           -- display best classification result\n"
			+ "                       --reset                                -- reset stats\n"
			+ "                       --version                              -- display version info\n"
			+ "                       --help | -h                            -- display this help and exit\n\n" +

			"Options (one or more of the following):\n\n" +

			"Loaders:\n\n"
			+ "  -wav          - assume WAVE files loading (default)\n"
			+ "  -text         - assume loading of text samples\n"
			+ "  -tiff         - assume loading of TIFF samples\n"
			+ "\n" +

			"Preprocessing:\n\n"
			+ "  -silence      - remove silence (can be combined with any of the below)\n"
			+ "  -noise        - remove noise (can be combined with any of the below)\n"
			+ "  -raw          - no preprocessing\n"
			+ "  -norm         - use just normalization, no filtering\n"
			+ "  -low          - use low-pass FFT filter\n"
			+ "  -high         - use high-pass FFT filter\n"
			+ "  -boost        - use high-frequency-boost FFT preprocessor\n"
			+ "  -band         - use band-pass FFT filter\n"
			+ "  -bandstop     - use band-stop FFT filter\n"
			+ "  -endp         - use endpointing\n"
			+ "  -lowcfe       - use low-pass CFE filter\n"
			+ "  -highcfe      - use high-pass CFE filter\n"
			+ "  -bandcfe      - use band-pass CFE filter\n"
			+ "  -bandstopcfe  - use band-stop CFE filter\n"
			+ "\n" +

			"Feature Extraction:\n\n"
			+ "  -lpc          - use LPC\n"
			+ "  -fft          - use FFT\n"
			+ "  -minmax       - use Min/Max Amplitudes\n"
			+ "  -randfe       - use random feature extraction\n"
			+ "  -aggr         - use aggregated FFT+LPC feature extraction\n"
			+ "  -f0           - use F0 (pitch, or fundamental frequency; NOT IMPLEMENTED\n"
			+ "  -segm         - use Segmentation (NOT IMPLEMENTED)\n"
			+ "  -cepstral     - use Cepstral analysis (NOT IMPLEMENTED)\n"
			+ "\n" +

			"Classification:\n\n"
			+ "  -nn           - use Neural Network\n"
			+ "  -cheb         - use Chebyshev Distance\n"
			+ "  -eucl         - use Euclidean Distance\n"
			+ "  -mink         - use Minkowski Distance\n"
			+ "  -diff         - use Diff-Distance\n"
			+ "  -zipf         - use Zipf's Law-based classifier\n"
			+ "  -randcl       - use random classification\n"
			+ "  -markov       - use Hidden Markov Models (NOT IMPLEMENTED)\n"
			+ "  -hamming      - use Hamming Distance\n"
			+ "  -cos          - use Cosine Similarity Measure\n"
			+ "\n" +

			"NLP/Ngrams/Smoothing:\n\n"
			+ "  -interactive  - interactive mode for classification instead of reading from a file\n"
			+ "  -char         - use characters as n-grams (should always be present for this app)\n\n"
			+ "  -unigram      - use UNIGRAM model\n"
			+ "  -bigram       - use BIGRAM model\n"
			+ "  -trigram      - use TRIGRAM model\n\n"
			+ "  -mle          - use MLE\n"
			+ "  -add-one      - use Add-One smoothing\n"
			+ "  -add-delta    - use Add-Delta (ELE, d=0.5) smoothing\n"
			+ "  -witten-bell  - use Witten-Bell smoothing\n"
			+ "  -good-turing  - use Good-Turing smoothing\n" +

			"\n"
			+ "  -spectrogram  - dump spectrogram image after feature extraction\n"
			+ "  -graph        - dump wave graph before preprocessing and after feature extraction\n"
			+ "  <integer>     - expected subject ID\n"
			+ "\n"
		);

		System.exit(0);
	}

	/**
	 * Retrieves String representation of the application's version.
	 * 
	 * @return version String
	 */
	public static final String getVersion()
	{
		return MAJOR_VERSION + "." + MINOR_VERSION + "." + REVISION;
	}

	/**
	 * Retrieves integer representation of the application's version.
	 * 
	 * @return integer version
	 */
	public static final int getIntVersion()
	{
		return MAJOR_VERSION * 100 + MINOR_VERSION * 10 + REVISION;
	}

	/**
	 * Makes sure the applications isn't run against older MARF version. Exits
	 * with 1 if the MARF version is too old.
	 */
	public static final void validateVersions()
	{
		if(MARF.getDoubleVersion() < (0 * 100 + 3 * 10 + 0 + .6))
		{
			System.err.println
			(
				"Your MARF version (" + MARF.getVersion()
				+ ") is too old. This application requires 0.3.0.6 or above."
			);

			System.exit(1);
		}
	}

	/**
	 * Composes the current configuration of in a string form.
	 * 
	 * @param pstrArgv
	 *            set of configuration options passed through the command line;
	 *            can be null or empty. If latter is the case, MARF itself is
	 *            queried for its numerical set up inside.
	 * 
	 * @return the current configuration setup
	 */
	public static final String getConfigString(String[] pstrArgv)
	{
		// Store config and error/successes for that config
		String strConfig = "";

		if(pstrArgv != null && pstrArgv.length > 2)
		{
			// Get config from the command line
			for(int i = 2; i < pstrArgv.length; i++)
			{
				// We do NOT want numerical IDs
				try
				{
					Integer.parseInt(pstrArgv[i]);
				}
				catch(NumberFormatException e)
				{
					// Skip .xml file names
					if(pstrArgv[i].matches(".*\\.xml"))
					{
						continue;
					}

					strConfig += pstrArgv[i] + " ";
				}
			}
		}
		else
		{
			// Query MARF for it's current config
			strConfig = MARF.getConfig();
		}

		return strConfig;
	}

	/**
	 * Sets default MARF configuration parameters as normalization for
	 * preprocessing, FFT for feature extraction, Euclidean distance for
	 * training and classification with no spectrogram dumps and no debug
	 * information, assuming WAVE file format.
	 * 
	 * @throws MARFException
	 */
	public static final void setDefaultConfig() throws MARFException
	{
		/*
		 * Default MARF setup
		 */
		MARF.setPreprocessingMethod(MARF.DUMMY);
		MARF.setFeatureExtractionMethod(MARF.FFT);
		MARF.setClassificationMethod(MARF.EUCLIDEAN_DISTANCE);
		MARF.setDumpSpectrogram(false);
		// MARF.setSampleFormat(MARF.WAV);
		// MARF.setSampleFormat(MARF.TEXT);
		MARF.setSampleFormat(MARF.CUSTOM);
		MARF.setSampleLoaderPluginClass(AnyToWAVLoader.class.getName());

		//Debug.enableDebug(false);
	}

	/**
	 * Customizes MARF's configuration based on the options.
	 * 
	 * @throws MARFException
	 *             if some options are out of range
	 */
	public static final void setCustomConfig()
	throws MARFException
	{
		ModuleParams oParams = new ModuleParams();

		for
		(
			int iPreprocessingMethod = MARF.MIN_PREPROCESSING_METHOD;
			iPreprocessingMethod <= MARF.MAX_PREPROCESSING_METHOD;
			iPreprocessingMethod++
		)
		{
			if(soGetOpt.isActiveOption(iPreprocessingMethod))
			{
				MARF.setPreprocessingMethod(iPreprocessingMethod);

				switch(iPreprocessingMethod)
				{
					// Endpointing did not respond well to silence removal
					// and the definition of RAW assumes none of that is done.
					// Endpointing also has some extra params of its own.
					case MARF.RAW:
					case MARF.ENDPOINT:

					case MARF.DUMMY:
					case MARF.HIGH_FREQUENCY_BOOST_FFT_FILTER:
					case MARF.HIGH_PASS_FFT_FILTER:
					case MARF.LOW_PASS_FFT_FILTER:
					case MARF.BANDPASS_FFT_FILTER:
					case MARF.BAND_STOP_FFT_FILTER:
					case MARF.HIGH_PASS_BOOST_FILTER:
					case MARF.BAND_PASS_CFE_FILTER:
					case MARF.BAND_STOP_CFE_FILTER:
					case MARF.LOW_PASS_CFE_FILTER:
					case MARF.HIGH_PASS_CFE_FILTER:
					{
						// Kludge
						if(soGetOpt.isActiveOption(OPT_SILENCE_NOISE_REMOVAL))
						{
							oParams.addPreprocessingParam(true);
							oParams.addPreprocessingParam(true);
						}
						else
						{
							// Normalization and filters seem to respond better
							// to silence removal.
							// The setting of the third protocol parameter
							// (silence threshold)
							// is yet to be implemented here.
							oParams.addPreprocessingParam(soGetOpt.isActiveOption(OPT_NOISE_REMOVAL));
							oParams.addPreprocessingParam(soGetOpt.isActiveOption(OPT_SILENCE_REMOVAL));
						}

						break;
					}

					default:
					{
						assert false : "Not implemented valid preprocessing configuration parameter: " + iPreprocessingMethod;
					}
				} // switch

				break;
			}
		}

		for
		(
			int iFeatureExtractionMethod = MARF.MIN_FEATUREEXTRACTION_METHOD;
			iFeatureExtractionMethod <= MARF.MAX_FEATUREEXTRACTION_METHOD;
			iFeatureExtractionMethod++
		)
		{
			if(soGetOpt.isActiveOption(iFeatureExtractionMethod))
			{
				MARF.setFeatureExtractionMethod(iFeatureExtractionMethod);

				switch(iFeatureExtractionMethod)
				{
					case MARF.FFT:
					case MARF.LPC:
					case MARF.RANDOM_FEATURE_EXTRACTION:
					case MARF.MIN_MAX_AMPLITUDES:
					case MARF.F0:
					case MARF.CEPSTRAL:
					case MARF.SEGMENTATION:
						// For now do nothing; customize when these methods
						// become parameterizable.
						break;

					case MARF.FEATURE_EXTRACTION_AGGREGATOR:
					{
						// For now aggregate FFT followed by LPC until
						// it becomes customizable
						oParams.addFeatureExtractionParam(new Integer(MARF.FFT));
						oParams.addFeatureExtractionParam(null);
						oParams.addFeatureExtractionParam(new Integer(MARF.LPC));
						oParams.addFeatureExtractionParam(null);
						break;
					}

					default:
						assert false;
				} // switch

				break;
			}
		}

		for
		(
			int iClassificationMethod = MARF.MIN_CLASSIFICATION_METHOD;
			iClassificationMethod <= MARF.MAX_CLASSIFICATION_METHOD;
			iClassificationMethod++
		)
		{
			if(soGetOpt.isActiveOption(iClassificationMethod))
			{
				MARF.setClassificationMethod(iClassificationMethod);

				switch(iClassificationMethod)
				{
					case MARF.NEURAL_NETWORK:
					{
						// Dump/Restore Format of the TrainingSet
						oParams.addClassificationParam(TrainingSet.DUMP_GZIP_BINARY);

						// Training Constant
						oParams.addClassificationParam(0.5);

						// Epoch number
						oParams.addClassificationParam(20);

						// Min. error
						oParams.addClassificationParam(0.1);

						break;
					}

					case MARF.HAMMING_DISTANCE:
					{
						// Dump/Restore Format
						oParams.addClassificationParam(TrainingSet.DUMP_GZIP_BINARY);

						// Type of hamming comparison
						oParams.addClassificationParam(HammingDistance.STRICT_DOUBLE);

						break;
					}

					case MARF.MINKOWSKI_DISTANCE:
					{
						// Dump/Restore Format
						oParams.addClassificationParam(TrainingSet.DUMP_GZIP_BINARY);

						// Minkowski Factor
						oParams.addClassificationParam(6.0);

						break;
					}

					case MARF.EUCLIDEAN_DISTANCE:
					case MARF.CHEBYSHEV_DISTANCE:
					case MARF.MAHALANOBIS_DISTANCE:
					case MARF.RANDOM_CLASSIFICATION:
					case MARF.DIFF_DISTANCE:
					case MARF.MARKOV:
					case MARF.ZIPFS_LAW:
					case MARF.COSINE_SIMILARITY_MEASURE:
					{
						// Dump/Restore Format
						oParams.addClassificationParam(TrainingSet.DUMP_GZIP_BINARY);

						// For now do nothing; customize when these methods
						// become parameterizable.
						break;
					}

					default:
						assert false : "Unrecognized classification module";
				} // switch

				// Method is found, break out of the look up loop
				break;
			}
		}

		// Assign meaningful params only
		if(oParams.size() > 0)
		{
			MARF.setModuleParams(oParams);
		}
	}
}

// EOF
\end{Verbatim}
\normalsize

\subsection{\api{PPMDumper}}
\scriptsize
\begin{Verbatim}
package marf.apps.oct.OCTMARF.Storage;

import java.io.BufferedOutputStream;
import java.io.DataOutputStream;
import java.io.FileOutputStream;
import java.util.Vector;

import marf.Storage.StorageException;
import marf.util.Debug;


/**
 * <p>Dumps an image to a PPM file.</p>
 *
 * @author Serguei Mokhov
 * @version $Id: PPMDumper.java,v 1.5 2012/01/02 08:06:33 mokhov Exp $
 * @since 0.0.1
 */
public class PPMDumper
{
	/**
	 * The data vector.
	 */
	protected Vector<Double[]> oData = null;

	/**
	 * Current minimum.
	 */
	protected double dMin = 0.0;

	/**
	 * Current maximum.
	 */
	protected double dMax = 0.0;

	/**
	 * To differentiate file names based on the feature extraction method name.
	 */
	protected String strMethod = "";

	/**
	 * Constructor.
	 */
	public PPMDumper()
	{
		this.oData = new Vector<Double[]>();
	}

	/**
	 * Constructor with a feature extraction method name.
	 * @param pstrMethodName String representing FE module name
	 */
	public PPMDumper(String pstrMethodName)
	{
		this();
		this.strMethod = pstrMethodName;
	}

	/**
	 * Dumps image.
	 * @throws StorageException
	 */
	public final void dump()
	throws StorageException
	{
		try
		{
			Debug.debug("Dumping image " + this.strMethod + ".ppm");
			Debug.debug(".dump() - data size in vectors: " + this.oData.size());

			FileOutputStream oFOS = null;
			DataOutputStream oOutFile = null;

//			oFOS = new FileOutputStream("58." + this.strMethod + ".ppm");
			//oFOS = new FileOutputStream("58.ppm");
			oFOS = new FileOutputStream(this.strMethod + ".ppm");
			//oOutFile = new DataOutputStream(oFOS);
			oOutFile = new DataOutputStream(new BufferedOutputStream(oFOS));
			
		// Output PPM header
/*
		man ppm:

         - A "magic number" for identifying the file type.  A pgm file's magic number is the two characters "P6".

         - Whitespace (blanks, TABs, CRs, LFs).

         - A width, formatted as ASCII characters in decimal.

         - Whitespace.

         - A height, again in ASCII decimal.

         - Whitespace.

         - The maximum color value (Maxval), again in ASCII decimal.  Must be less than 65536.

         - Newline or other single whitespace character.

         - A raster of Width * Height pixels, proceeding through the image in normal English reading order.  Each pixel is  a  triplet  of  red,
           green,  and  blue  samples,  in that order.  Each sample is represented in pure binary by either 1 or 2 bytes.  If the Maxval is less
           than 256, it is 1 byte.  Otherwise, it is 2 bytes.  The most significant byte is first.
*/
			// Output data

			// Make max be at 75%
			this.dMax = 255;

			int iWidth = this.oData.size();
			int iHeight = this.oData.elementAt(0).length;

			oOutFile.writeBytes
			(
				"P6\n"
				+ iWidth + "\n"
				+ iHeight + "\n"
				+ "255\n"
			);

			for(int i = iHeight - 1; i >= 0; i--)
			{
				for(int j = 0; j < iWidth; j++)
				{
					Double[] adData = this.oData.elementAt(j);

					// Colors, RGB, hence 3
 					for(int m = 0; m < 3; m++)
					{
 						oOutFile.writeByte(adData[i].byteValue());
					}
				}
			}
			
			oOutFile.flush();

			Debug.debug
			(
				"Done dumping image " +
				this.strMethod +
				".ppm [" + (this.oData.size() * iHeight * iWidth) + " bytes]"
			);
		}
		catch(Exception e)
		{
			throw new StorageException(e);
		}
	}

	public void addToData(double[] padData)
	{
		Double[] oNewData = new Double[padData.length];
		//Arrays.copy(oNewData, paData);
		
		for(int i = 0; i < padData.length; i++)
		{
			oNewData[i] = padData[i];
		}
		
		this.oData.add(oNewData);
	}

	public void addToData(Double[] paData)
	{
		this.oData.add(paData);
	}

	/**
	 * @return the oData
	 */
	public Vector<Double[]> getData()
	{
		return this.oData;
	}

	/**
	 * @param oData the oData to set
	 */
	public void setData(Vector<Double[]> poData)
	{
		this.oData = poData;
	}
	
	public String getFilename()
	{
		return this.strMethod;
	}
}

// EOF
\end{Verbatim}
\normalsize

\addcontentsline{toc}{section}{Index}
\printindex

\end{document}